\documentclass{article}
\usepackage{arxiv}
\usepackage{amsmath,epsfig, caption, subcaption, float}

\title{Transfer Learning of Semantic Segmentation Methods for Identifying Buried Archaeological Structures on LiDAR Data}

\author{
 Gregory Sech \\
  Center for Cultural Heritage Technology\\
  Istituto Italiano di Tecnologia \\
  Venice, Italy \\
   \And
 Paolo Soleni \\
  Center for Cultural Heritage Technology\\
  Istituto Italiano di Tecnologia \\
  Venice, Italy \\
  \And
 Wouter B. Verschoof-van der Vaart \\
  Leiden University\\
  Faculty of Archaeology\\
  Leiden, Netherlands\\
\And
 \v{Z}iga Kokalj \\
 Research Centre of the Slovenian \\ Academy of Sciences and Arts\\
 Ljubljana, Slovenia\\
\And
 Arianna Traviglia \\
  Center for Cultural Heritage Technology\\
  Istituto Italiano di Tecnologia \\
  Venice, Italy \\
\And
 Marco Fiorucci \\
  Center for Cultural Heritage Technology\\
  Istituto Italiano di Tecnologia \\
  Venice, Italy \\
  Kyoto University\\
  Kyoto, Japan\\
}  

\begin{document}

\maketitle
\begin{abstract}
When applying deep learning to remote sensing data in archaeological research, a notable obstacle is the limited availability of suitable datasets for training models. The application of transfer learning is frequently employed to mitigate this drawback. However, there is still a need to explore its effectiveness when applied across different archaeological datasets. This paper compares the performance of various transfer learning configurations using two semantic segmentation deep neural networks on two LiDAR datasets. The experimental results indicate that transfer learning-based approaches in archaeology can lead to performance improvements, although a systematic enhancement has not yet been observed. We provide specific insights about the validity of such techniques that can serve as a baseline for future works.
\end{abstract}
\section{Introduction}
\label{sec:intro}
The use of Deep Learning (DL) based workflows in archaeological research has become widespread in the last decade \cite{bickler_2021_advances_in_achaeological_practice} as a result of an increasing interest in the automatic detection of buried archaeological structures in remote sensing imagery, in turn, encouraged by the ever-growing amount of high-quality data collected by airborne and spaceborne sensors. 
One of the main shortcomings in developing DL approaches for archaeology is the limited availability of suitable-sized labelled datasets that are required to train deep neural networks due to the expertise and time needed to annotate them \cite{fiorucci_2020_pattern_recognition_letters}. Transfer Learning (TL), i.e. the use of models pre-trained on general datasets (e.g., ImageNet \cite{deng_2009_cvpr}), has been extensively employed in archaeology to overcome this issue \cite{verschoof_2022_thesis}.
While TL between general datasets and satellite imagery has already been studied \cite{pires_2020_remote_sensing}, the effectiveness of TL applied across different archaeological LiDAR datasets remains to be explored \cite{gallwey_2019_remote_sensing}. This paper aims to fill this gap by presenting the first systematic experimental study about the effectiveness of TL in archaeological research. We compare the performances of several configurations of two DL architectures for semantic segmentation on two different archaeological LiDAR datasets. As a result, we also obtain the benchmark of two semantic segmentation networks for identifying archaeological objects, which can serve as a baseline for future algorithm development. 
The main implications of this study for the archaeological community are discussed, providing specific insights into the ability of pre-trained networks to capture the shapes and textures of archaeological structures.

\section{Datasets} 
Two LiDAR datasets--diverging in terms of represented geological, topographic, climatic, vegetational, and historical features, as well as in terms of observed archaeological remains--have been used in this research. They consist of tiles ($256 \times 256$ pixels) realised with Enhanced Multiscale Topographic Position (e2MSTP) and contain three labelled classes corresponding to ancient human-made structures. The Chact\'un dataset \cite{somrak_2020_remote_sensing} ($3868$ tiles of which 10 empty tiles, i.e. without archaeological objects) includes aguadas (artificial water reservoirs), buildings, and platforms from central Yucatan, Mexico, while the Veluwe dataset \cite{fiorucci_2022_remote_sensing} ($3539$ tiles with $20$ empty tiles) includes barrows (round earthen mounds demarcating burials), Celtic fields (checkerboard-patterned parcelling system of contiguous embanked plots), and charcoal kilns (circular platforms surrounded by a shallow ditch for the production of charcoal) from the Netherlands. Both datasets present a large class imbalance (Table  \ref{table:1}).

\begin{table}[ht]
\centering
\caption{Overview and distribution of labelled classes in the Chact\'un and Veluwe datasets.  
Both datasets exhibit significant class imbalance and include empty tiles. 
A red segmentation mask corresponding to the labelled pixels is overlayed on the raster image to highlight the respective class.\\}
\label{table:1}
\begin{tabular}{|l|l|l|l|}

\hline
Chact\'un & aguadas & buildings & platforms \\
\hline
Nr. of tiles & 166 & 3532 & 2472\\
\hline
                & \includegraphics[width=0.09\textwidth]{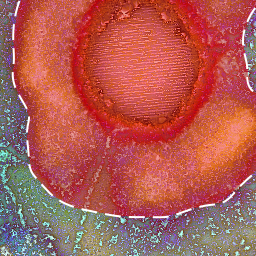}
                & \includegraphics[width=0.09\textwidth]{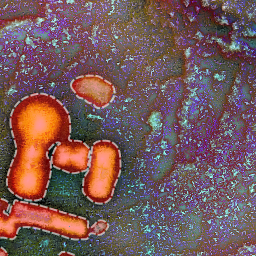}             
                & \includegraphics[width=0.09\textwidth]{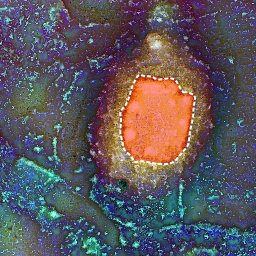} \\
\hline
Veluwe          & barrows & Celtic fields & charcoal kilns\\
\hline
Nr. of tiles & 993 & 2365 & 332\\
\hline
                & \includegraphics[width=0.09\textwidth]{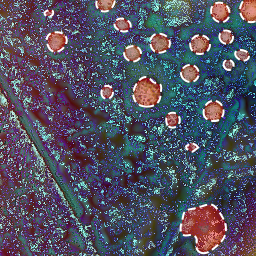}         
                & \includegraphics[width=0.09\textwidth]{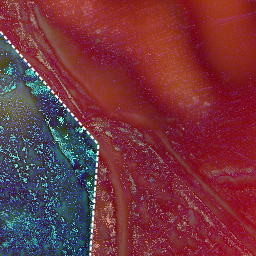}             
                & \includegraphics[width=0.09\textwidth]{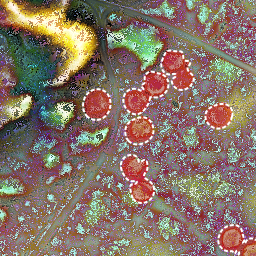}\\
\hline
\end{tabular}
\end{table}

\section{Methodology}
In this work, we characterise how TL can impact neural networks' capability to identify archaeological landscape objects on remote sensing data. To this end, we designed a comparative study of two TL scenarios: in the first, we selected a computer vision model already pre-trained on a commonly adopted benchmark dataset (ImageNet) and we trained it on an archaeological dataset. As an additional baseline, we also considered models trained from scratch using Kaiming weight initialisation \cite{he_2015_iccv}, a sound initialisation method addressing rectifier nonlinearities and weights convergence that has proven effective with ReLU-based networks. In the second scenario, we employed TL on the models pre-trained on one archaeological dataset and applied them to the other.
To assess the effectiveness of the proposed scenarios, we compared them via the performance of two popular semantic segmentation architectures, namely U-Net \cite{ronneberger_2015_miccai2015} and DeepLabV3+ \cite{chen_2018_eccv}, using different backbone feature extractor configurations: ResNet \cite{he_2015_cvpr}  and EfficientNet \cite{tan_2019_icml} for both, and additionally, SegFormer \cite{xie_2021_neurips} for U-Net. 
\subsection{Semantic Segmentation} %
DL-based object detection methods are used in archaeology for classifying and localising archaeological objects \cite{verschoof_2022_thesis}. Despite being a valuable technique, object detection exhibits certain limitations, such as the lack of information regarding the shapes of the identified objects which is a significant drawback.
In contrast, semantic segmentation provides a more detailed and nuanced understanding of an image, allowing an enhanced comprehension of the exact location and the shapes of the objects. This can be particularly useful in archaeology, as detailed information on the scale and geospatial extent of archaeological objects can be important for an improved understanding of the significance of a site.
Among the state-of-the-art methods, we selected U-Net\cite{ronneberger_2015_miccai2015} and DeepLabV3+\cite{chen_2018_eccv}.
Both introduced several technical solutions to address some of the main challenges in the field, such as enhancing the combination of high-level context and detailed spatial information. The first presented a symmetric expansive upsampling path and skip connections, while the second is the latest of a series of architectures that progressively proposed several novelties based on atrous convolutions and spatial pyramid pooling.

\section{Experiments}
To assess the generalisation capability of a transfer-learned model in archaeology, the five models were trained in two different situations, i.e., a general computer vision TL and an archaeological TL scenario. In the former, the initial network weights were obtained with the Kaiming technique or by pre-training on ImageNet, and subsequently, the models were fine-tuned on each of the two archaeological target datasets. In the latter, trained models from the first scenario were further fine-tuned on the complementary target archaeological dataset. In this way, we have been able to assess whether archaeological features are learned from the tuning even though a large archaeological dataset is unavailable. A 5-fold cross-validation was used for each model in both scenarios to better harden our empirical results against statistical fluctuations.
To measure model performance, we employed mean Intersection over Union (mIoU), which is obtained by averaging Intersection over Union (IoU) on both the tiles and the dataset's classes.
To fairly compare the scenarios, we conducted an automatic hyperparameter optimisation (HPO) search of each network architecture. Ray Tune's 
\cite{liaw_2018_arxiv}
implementation of the Asynchronous HyperBand scheduling \cite{li_2020_mlsys} was used to tune our optimiser's (Adam \cite{kingma2014adam}) parameters as well as to choose the most appropriate batch size.

\begin{figure*}[t]
     \centering
     \begin{subfigure}[b]{0.48\textwidth}
         \centering
         \caption{target dataset: Veluwe}
         \label{fig:miou_veluwe}
         \includegraphics[width=\textwidth, height=3in]{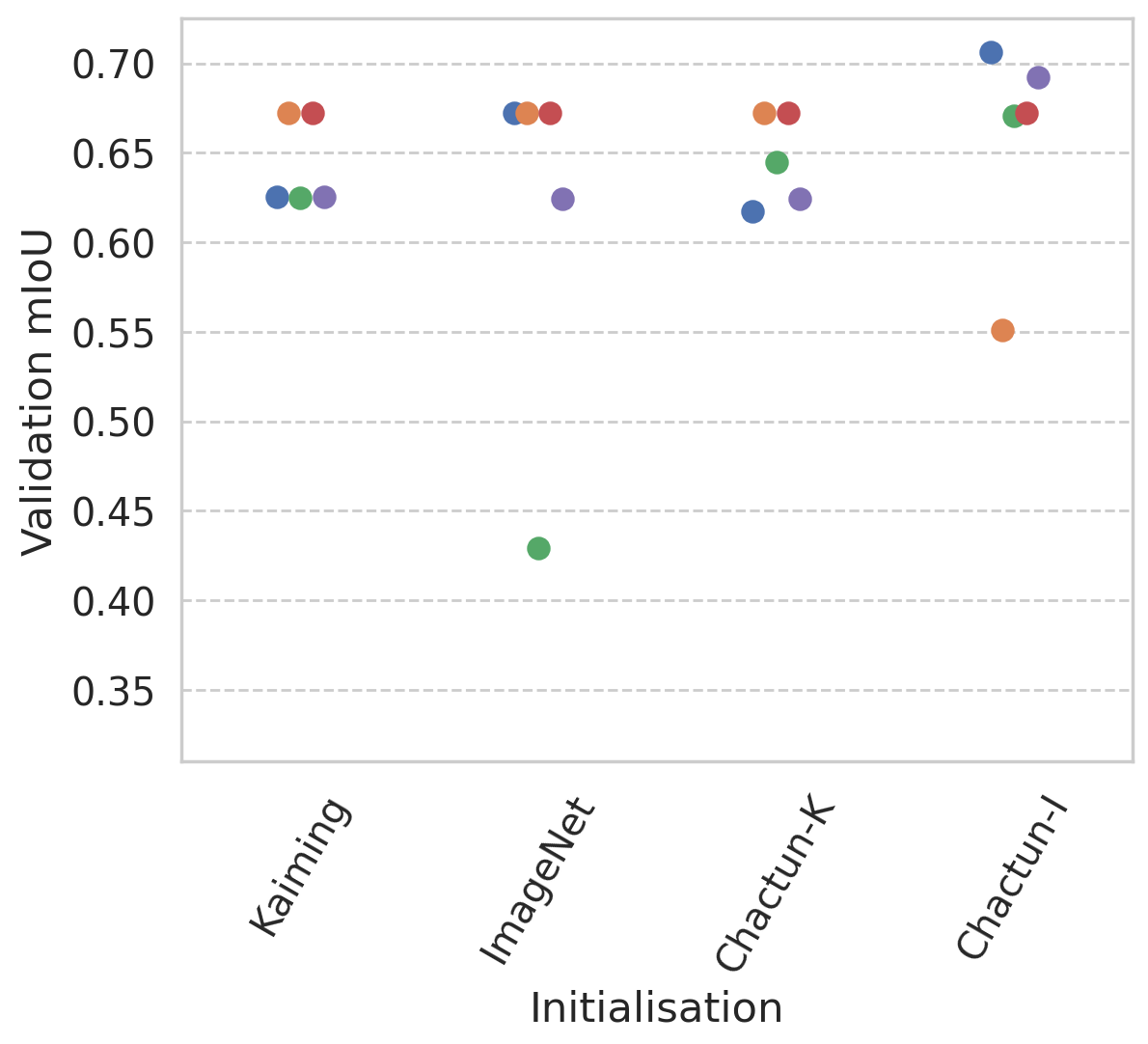}
     \end{subfigure}
     \begin{subfigure}[b]{0.48\textwidth}
         \centering
         \caption{target dataset: Chact\'un}
         \label{fig:miou_chactun}
         \includegraphics[width=\textwidth, height=3in]{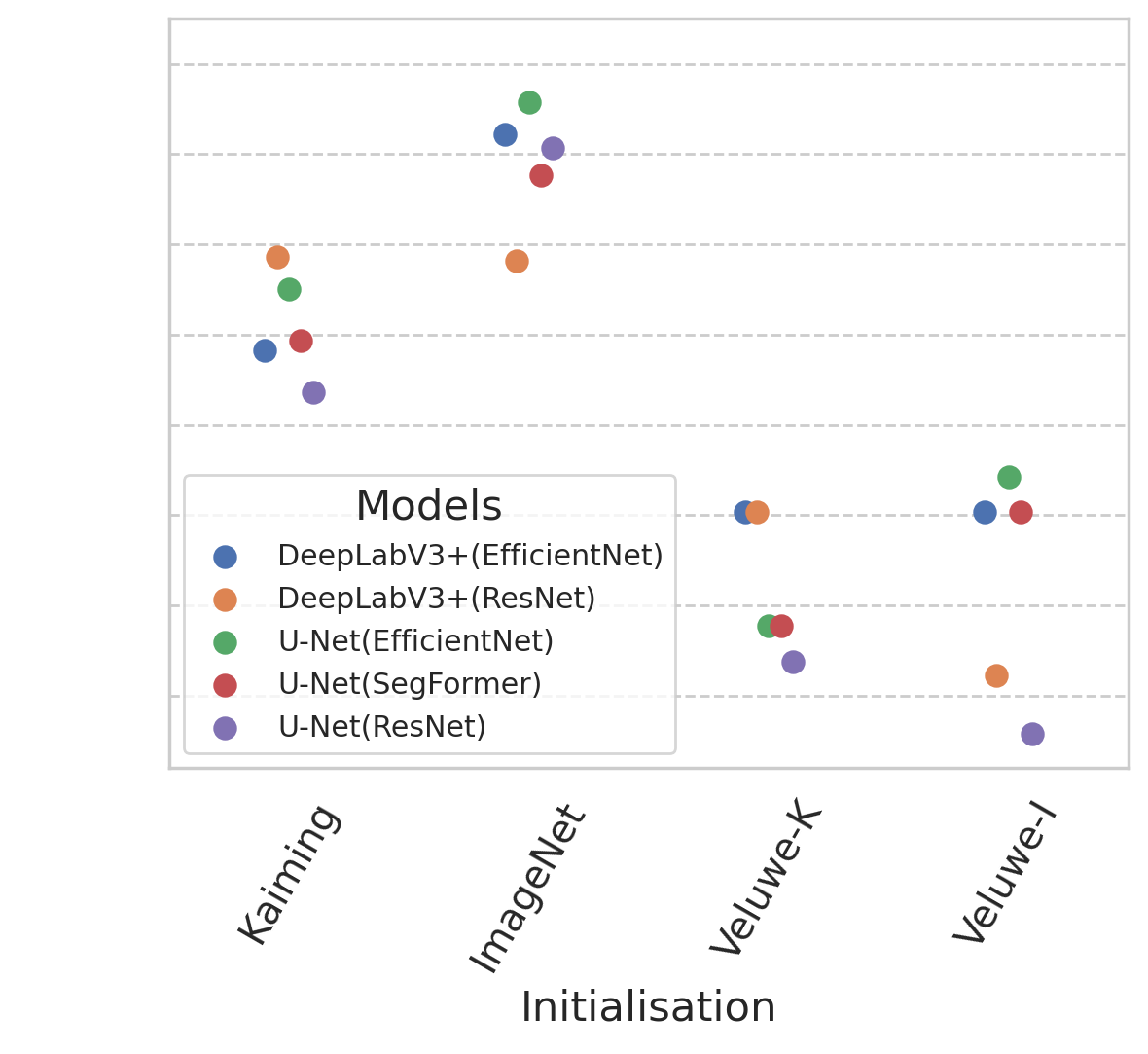}
     \end{subfigure}
     \caption{Impact of weight initialisation on mIoU of model configurations on target datasets. mIoU is estimated using cross-validation. Model configurations are U-Net employing backbone feature extractors (ResNet, EfficientNet, SegFormer) and DeepLabV3+ (ResNet and EfficientNet). Four weights initialisations are considered: no pre-training (Kaiming), TL from ImageNet (ImageNet), TL from the other archaeological dataset of a model with no pre-training (\{Veluwe,Chact\'un\}-K), TL from the other archaeological dataset of a model with ImageNet pre-training (\{Veluwe,Chact\'un\}-I).} 
     \label{fig:miou_initialisations}
\end{figure*}

\subsection{Results}
The cross-validated mIoU for each dataset and weight initialisation was analysed to assess the contribution of
TL. The best performance (up to 5\%) on the Veluwe target dataset was achieved by one transfer-learned method (namely, the Chact\'un-I initialisation), while all other TL experiments, namely Chact\'un-K and ImageNet, did not show any systematic improvement on this dataset with respect to the Kaiming initialisation without pre-training (Fig.\ref{fig:miou_veluwe}). On the Chact\'un target dataset, the only positive contribution of TL was observed when the ImageNet pre-training was used: it systematically outperformed all the other experiments, achieving an improvement of 11\% over Kaiming. The other TL attempts (i.e., Veluwe-K and Veluwe-I) have a remarkable negative impact on the final performance since even the Kaiming initialised models performed better than them (Fig.\ref{fig:miou_chactun}).

\section{Discussion}
The study revealed that the two TL scenarios produced mixed results. In general, TL demonstrated its effectiveness by providing a more convenient starting point for model initialisation than training from scratch. Pre-training on natural images resulted in similar or even better performance on both target archaeological datasets compared to no pre-training (Kaiming) with minimal additional cost. On the other hand, pre-training on archaeological datasets did not consistently yield performance improvements.  
Transfer-learned Veluwe models, when subsequently fine-tuned on the Chact\'un dataset, exhibited a detrimental effect, whereas the opposite was observed in the complementary case, where a performance improvement was observed. This contradiction can be attributed to dissimilarities between the two datasets. Given the experimental set-up, variations in image size, LiDAR data quality, and visualisation, which typically contribute to performance differences can be ruled out \cite{guyot_2021_journal_of_archaeological_science} \cite{verschooflandauer_2022}. Experimental evidence suggests that the distinctive characteristics of each dataset class (see Table \ref{table:1}) also influence the final performance of the model during fine-tuning \cite{Dolesj2020}. We suppose that the elementary features learned from Veluwe, which predominantly consist of relatively small circular shapes representing barrows and charcoal kilns, inhibit the extraction of additional features that characterise objects in the Chact\'un dataset. This dataset primarily comprises relatively large, squared-off shapes for buildings and platforms. In addition, the classes in the Veluwe dataset primarily exhibit positive elevations (protruding from the ground), while the Chact\'un dataset includes classes with negative elevations (i.e., aguadas).

\section{Conclusion}
The experiments demonstrated that TL is a valuable tool for building an effective DL-based workflow for the identification of archaeological structures. Diverse TL scenarios provide mixed results, likely due to differences in the inherent characteristics of the archaeological objects within the two considered datasets. Based on these considerations, we argue that the construction of a large new dataset featuring a wide variety of heterogeneous archaeological objects will encourage the effective use of TL between archaeological datasets.

\section*{Acknowledgment}
This project 	, and partially funded by the Slovenian Research Agency's research program Earth observation and geoinformatics (P2-0406). This research was also funded by the European Union’s Horizon 2020 research and innovation programme under grant agreement No 101027956. We gratefully acknowledge the HPC infrastructure and the Support Team at the Istituto Italiano di Tecnologia.

\bibliographystyle{IEEEbib}
\bibliography{refs}

\begin{thebibliography}{10}

\bibitem{bickler_2021_advances_in_achaeological_practice}
Simon~H. Bickler,
\newblock ``Machine learning arrives in archaeology,''
\newblock {\em Advances in Archaeological Practice}, 2021.

\bibitem{fiorucci_2020_pattern_recognition_letters}
Marco Fiorucci, Marina Khoroshiltseva, Massimiliano Pontil, Arianna Traviglia,
  Alessio {Del Bue}, and Stuart James,
\newblock ``Machine learning for cultural heritage: A survey,''
\newblock {\em Pattern Recognition Letters}, 2020.

\bibitem{deng_2009_cvpr}
Jia Deng, Wei Dong, Richard Socher, Li-Jia Li, Kai Li, and Li~Fei-Fei,
\newblock ``Imagenet: A large-scale hierarchical image database,''
\newblock in {\em 2009 IEEE Conference on Computer Vision and Pattern
  Recognition}, 2009.

\bibitem{verschoof_2022_thesis}
Wouter {Verschoof-van der Vaart},
\newblock {\em Learning to look at LiDAR: combining CNN-based object detection
  and GIS for archaeological prospection in remotely-sensed data},
\newblock Ph.D. thesis, Leiden University, 2022.

\bibitem{pires_2020_remote_sensing}
Rafael Pires~de Lima and Kurt Marfurt,
\newblock ``Convolutional neural network for remote-sensing scene
  classification: Transfer learning analysis,''
\newblock {\em Remote Sensing}, vol. 12, 2020.

\bibitem{gallwey_2019_remote_sensing}
Jane Gallwey, Matthew Eyre, Matthew Tonkins, and John Coggan,
\newblock ``Bringing lunar lidar back down to earth: Mapping our industrial
  heritage through deep transfer learning,''
\newblock {\em Remote Sensing}, vol. 11, 2019.

\bibitem{somrak_2020_remote_sensing}
Maja Somrak, Sašo Džeroski, and \v{Z}iga Kokalj,
\newblock ``Learning to classify structures in als-derived visualizations of
  ancient maya settlements with cnn,''
\newblock {\em Remote Sensing}, 07 2020.

\bibitem{fiorucci_2022_remote_sensing}
Marco Fiorucci, Wouter Verschoof-van~der Vaart, Paolo Soleni, Bertrand Saux,
  and Arianna Traviglia,
\newblock ``Deep learning for archaeological object detection on lidar: New
  evaluation measures and insights,''
\newblock {\em Remote Sensing}, 03 2022.

\bibitem{he_2015_iccv}
Kaiming He, Xiangyu Zhang, Shaoqing Ren, and Jian Sun,
\newblock ``Delving deep into rectifiers: Surpassing human-level performance on
  imagenet classification,''
\newblock in {\em 2015 IEEE International Conference on Computer Vision
  (ICCV)}, 2015.

\bibitem{ronneberger_2015_miccai2015}
Olaf Ronneberger, Philipp Fischer, and Thomas Brox,
\newblock ``U-net: Convolutional networks for biomedical image segmentation,''
\newblock in {\em Medical Image Computing and Computer-Assisted Intervention --
  MICCAI 2015}, Nassir Navab, Joachim Hornegger, William~M. Wells, and
  Alejandro~F. Frangi, Eds. 2015, Springer International Publishing.

\bibitem{chen_2018_eccv}
Liang-Chieh Chen, Yukun Zhu, George Papandreou, Florian Schroff, and Hartwig
  Adam,
\newblock ``Encoder-decoder with atrous separable convolution for semantic
  image segmentation,''
\newblock in {\em Computer Vision – ECCV 2018: 15th European Conference,
  Munich, Germany, September 8-14, 2018, Proceedings, Part VII}. 2018,
  Springer-Verlag.

\bibitem{he_2015_cvpr}
Kaiming He, X.~Zhang, Shaoqing Ren, and Jian Sun,
\newblock ``Deep residual learning for image recognition,''
\newblock {\em 2016 IEEE Conference on Computer Vision and Pattern Recognition
  (CVPR)}, 2015.

\bibitem{tan_2019_icml}
Mingxing Tan and Quoc Le,
\newblock ``{E}fficient{N}et: Rethinking model scaling for convolutional neural
  networks,''
\newblock in {\em Proceedings of the 36th International Conference on Machine
  Learning}, Kamalika Chaudhuri and Ruslan Salakhutdinov, Eds. 09--15 Jun 2019,
  PMLR.

\bibitem{xie_2021_neurips}
Enze Xie, Wenhai Wang, Zhiding Yu, Anima Anandkumar, Jose~M. Alvarez, and Ping
  Luo,
\newblock ``Segformer: Simple and efficient design for semantic segmentation
  with transformers,''
\newblock in {\em Advances in Neural Information Processing Systems},
  M.~Ranzato, A.~Beygelzimer, Y.~Dauphin, P.S. Liang, and J.~Wortman Vaughan,
  Eds. 2021, vol.~34, Curran Associates, Inc.

\bibitem{liaw_2018_arxiv}
Richard Liaw, Eric Liang, Robert Nishihara, Philipp Moritz, Joseph~E. Gonzalez,
  and Ion Stoica,
\newblock ``Tune: A research platform for distributed model selection and
  training,'' 2018.

\bibitem{li_2020_mlsys}
Liam Li, Kevin Jamieson, Afshin Rostamizadeh, Ekaterina Gonina, Jonathan
  Ben-tzur, Moritz Hardt, Benjamin Recht, and Ameet Talwalkar,
\newblock ``A system for massively parallel hyperparameter tuning,''
\newblock in {\em Proceedings of Machine Learning and Systems}, I.~Dhillon,
  D.~Papailiopoulos, and V.~Sze, Eds., 2020, vol.~2.

\bibitem{kingma2014adam}
Diederik~P Kingma and Jimmy Ba,
\newblock ``Adam: A method for stochastic optimization,''
\newblock {\em arXiv preprint arXiv:1412.6980}, 2014.

\bibitem{guyot_2021_journal_of_archaeological_science}
Alexandre Guyot, Marc Lennon, and Laurence Hubert-Moy,
\newblock ``Objective comparison of relief visualization techniques with deep
  cnn for archaeology,''
\newblock {\em Journal of Archaeological Science: Reports}, 08 2021.

\bibitem{verschooflandauer_2022}
Wouter Verschoof-van~der Vaart and Juergen Landauer,
\newblock ``Testing the transferability of carcassonnet. a case study to detect
  hollow roads in germany and slovenia,''
\newblock in {\em Proceedings of the 25th Conference on Cultural Heritage and
  New Technologies, CHNT25}. 2022, Probylaeum, Heidelberg University Library.

\bibitem{Dolesj2020}
Martin Dolejš, Jan Pacina, Martin Veselý, and Dominik Brétt,
\newblock ``Aerial bombing crater identification: Exploitation of precise
  digital terrain models,''
\newblock {\em ISPRS International Journal of Geo-Information}, 2020.

\end{thebibliography}

\end{document}